\documentclass[pdflatex, sn-basic]{sn-jnl}

\usepackage{graphicx}%
\usepackage{multirow}%
\usepackage{amsmath,amssymb,amsfonts}%
\usepackage{amsthm}%
\usepackage{mathrsfs}%
\usepackage[title]{appendix}%
\usepackage{xcolor}%
\usepackage{textcomp}%
\usepackage{manyfoot}%
\usepackage{booktabs}%
\usepackage{algorithm}%
\usepackage{algorithmicx}%
\usepackage{algpseudocode}%
\usepackage{listings}%
\usepackage{indentfirst} 
\usepackage{enumitem} 
\usepackage{geometry} 
\usepackage[justification=centering]{caption}

\theoremstyle{thmstyleone}%
\theoremstyle{thmstyletwo}%
\newtheorem{example}{Example}%
\newtheorem{remark}{Remark}%

\theoremstyle{thmstylethree}%
\newtheorem{definition}{Definition}%

\begin{document}

\title{Ensemble Interpretation: A Unified Method for Interpretable Machine Learning}


\author*[1, 2, 3]{\fnm{Min} \sur{Chao}}\email{minchao@swpu.edu.cn}

\author[1, 2]{\fnm{Liao} \sur{Guoyong}}

\author[1, 2]{\fnm{Wen} \sur{Guoquan}}

\author[1, 2]{\fnm{Li} \sur{Yingjun}}

\author[1, 2]{\fnm{Guo} \sur{Xing}}

\affil*[1]{\orgdiv{School of Science}, \orgname{Southwest Petroleum University}, \orgaddress{\city{Chengdu}, \postcode{610500}, \country{China}}}

\affil[2]{\orgdiv{Institute for Artificial Intelligence}, \orgname{Southwest Petroleum University}, \orgaddress{\city{Chengdu}, \postcode{610500}, \country{China}}}

\affil[3]{\orgdiv{State Key Laboratory of Oil and Gas Reservoir and Exploitation}, \orgaddress{\city{Chengdu}, \postcode{610500}, \country{China}}}


\abstract{To address the issues of stability and fidelity in interpretable learning, a novel interpretable methodology, ensemble interpretation, is presented in this paper which integrates multi-perspective explanation of various interpretation methods. On one hand, we define a unified paradigm to describe the common mechanism of different interpretation methods, and then integrate the multiple interpretation results to achieve more stable explanation. On the other hand, a supervised evaluation method based on prior knowledge is proposed to evaluate the explaining performance of an interpretation method. The experiment results show that the ensemble interpretation is more stable and more consistent with human experience and cognition. As an application, we use the ensemble interpretation for feature selection, and then the generalization performance of the corresponding learning model is significantly improved.}




\maketitle

\section{Introduction}\label{sec1}
The interpretability of complex machine learning model is scarce because one cannot know 
how the predictions are made internally. In order to explain the prediction of machine learning model, 
interpretable machine learning is proposed to enable us  understanding the behavior of learning model and 
decide whether to trust the prediction results.
The generalization performance of machine learning model is getting much better, but in some high-risk industrial areas with insufficient samples\citep{schroff2015facenet,sun2015deepid3,taigman2014deepface,geiger2012we,tobiyama2016malware,rajpurkar2017chexnet}, the interpretability of these kind models are still inadequate. Researchers realize that it is unrealistic in industrial areas to pursue the generalization ability without interpretability \citep{rudin2022interpretable}. It has been reached an agreement that interpretability is a key factor in the credibility of industrial artificial intelligence \citep{rudin2022interpretable}.
At present, many methods have been developed to explain the output of a machine learning model, 
such as PDP \citep{friedman2001greedy}, ICE \citep{goldstein2015peeking}, LIME \citep{ribeiro2016should}, SHAP \citep{lundberg2017unified}, etc. However, these methods have some limitations in terms of fidelity and stability, 
and sometimes the explanation results of different methods may be conflicting \citep{garreau2020explaining, slack2020fooling, rahnama2019study}. 
For example, the random perturbation sampling method of LIME makes the generated samples scattered, resulting in some data not conforming to the original data distribution. Secondly, the samples generated by repeated experiments are also different, which leads to the lack of local fidelity and stability of LIME.

The interpretation of a black box model by a single interpretation method is like that 
a blind man gropes the elephant. Each method can only explain the black box from a partial perspective, 
but cannot demonstrate the whole structure of the model. 
Inspired by ensemble learning\citep{freund1996experiments,breiman1996bagging,alvear2018building,ren2016ensemble}, 
we consider applying the ensemble methodology of interpretability. 
In order to explain the black box model from multiple perspectives as much as possible, 
a method called ensemble interpretation, is presented in this paper, 
which integrates multiple interpretation methods to obtain a more comprehensive interpretation result. 
Specifically, we construct a mechanism to unify the interpretation methods into one paradigm. 
By defining composite mapping acting on this paradigm, 
the results of the interpretation methods can be mapped into a interpretation list. 
Finally, the ensemble interpretation result can be obtained by integrating the interpretation lists of multiple interpretation methods.

In this paper, a novel methodology for interpretable machine learning is presented 
based on the idea of ensemble interpretation. 
Section 1 is the introduction to the original intention of our research. 
In Section 2, the related works of interpretable machine learning are summarized. 
And in Section 3, a unified paradigm is constructed to express the mechanism of the interpretation method. 
The results of the interpretation method are also uniformly recorded as an interpretation list. 
In Section 4, different interpretation methods are integrated. 
And an index to evaluate the performance of the interpretation method is proposed. 
In Section 5, the experiment of ensemble interpretation is carried out, 
and the results show that the performance of the ensemble interpretation is better than 
that of the single interpretation method.

The main contributions of this paper are as follows: 
\\$\bullet$ The idea of ensembling are applied to interpretable machine learning, which can obtain more stable and multi-angle interpretation results.
\\$\bullet$ Interpretation list is presented to unify different interpretation methods into one paradigm, which describes the interpretation methods essentially.
\\$\bullet$ A supervised evaluation method is proposed to evaluate the quality of the Interpretation method.

\section{Related works}\label{sec2}
In general, interpretable methods can be categorized into two types: the ante-hoc interpretability and the post-hoc interpretability.

Ante-hoc interpretability means that the interpretability comes from the leaning model itself, 
which in general has a simple self-explanatory structure, including linear models and decision trees. 
In 2010, Štrumbelj et al. \citep{strumbelj2010efficient} proposed a general method for interpreting individual predictions of classification models based on basic concepts of game theory, 
and explained predictions based on the contributions of individual feature values. 
Ribeiro et al. \citep{goldstein2015peeking} proposed LIME in 2016, a model-agnositc local interpretation method and the main idea of this method is to construct a locally interpretable linear model near a specific sample for explaining the output of a complex model. 
Luo et al. \citep{lou2012intelligible} defined shape functions in generalized additive models (GAMs),
and combined the single feature models of shape functions using a linear model to explain the output of complex models, 
the authors also proposed a tree ensemble method based on adaptive leaf number, 
which was better than previous works. In decision tree models, each branch of the tree represents a decision result, 
and each path from the root node to different leaf nodes represents a decision rule.
Therefore, decision trees can be linearly represented as a series of decision rules composed of if-then statements \citep{huysmans2011empirical, breslow1997simplifying, quinlan1987generating}, 
which are very consistent with human cognition, making decision trees a self-explanatory model with strong interpretability.

Post-hoc interpretability provides explanations for a trained model, 
and the key is to establish an interpretation method with high local fidelity and stability to explain the black box model. 
Scott et al. \citep{friedman2001greedy} proposed a unified prediction explanation framework, SHAP (Shapley Additive Explanations). 
This method measures the influence of each feature on the output by calculating a Shapley value for each feature in relation to the prediction, 
this method has better computational performance and is more in line with human cognition than traditional methods. 
Later, Ribeiro et al. \citep{ribeiro2018anchors, ribeiro2016nothing} proposed a local explanation method called Anchor, 
which approximates the local boundary of a complex model using if-then rules and this method is easy to understand and very intuitive.
Partial Dependence Plots (PDP), proposed by Jerome et al. \citep{ribeiro2016should}, is a very advantageous visualization explanation tool, 
which visualizes the average relationship between the predicted response and one or more features. 
Individual Conditional Expectation (ICE), proposed by Alex et al. \citep{lundberg2017unified}, is also a visualization tool for interpretation methods that highlights the average relationship between the predicted variable and the predicted response. 
This method is an extension of the PDP method. Simonyan et al. \citep{simonyan2013deep} proposed the Grad method to characterize the importance of features for prediction through backpropagation, 
which calculates the gradient of the output with respect to the input by backpropagation to obtain a Saliency Map. 
DeconvNet, proposed by Zeiler et al. \citep{zeiler2014visualizing}, backpropagates the high-level activations of a deep neural network to the input of the model so that the parts responsible for activation in the input can be identified. 
Du et al. \citep{du2018towards} proposed an instance-level feature inversion explanation framework, 
which accurately locates the features in the input sample that affect the model output by adding category constraints during the guided feature inversion process. 
Zhou et al. \citep{zhou2016learning} proposed the Class Activation Mapping (CAM) explanation method, 
which replaces all fully connected layers in classical CNNs with a global average pooling layer, 
except for the softmax layer, and maps the weights of the output layer to the convolutional feature map to identify important regions in the image.

Above all, the connotation of interpretability is still not clear enough \citep{rudin2022interpretable}, a solid theoretical foundation for interpretable machine learning is in an imposing demand. 
Moreover, individual interpretation methods may suffer from insufficient stability \citep{rahnama2019study}. 
Thus we are trying find a novel thought to integrate various methods and enhance the stability of the interpretation results.

\section{The Unified Paradigm of Interpretation Methods}\label{sec3}
The essence of interpretation methods is to quantify the contribution of the input features to the output. 
Given a dataset $D$, suppose that $\varphi$ is a machine learning model trained from $D$. 
Let the input features and output of $\varphi$ be $X=\left\{x_{1}, x_{2}, \cdots, x_{n}\right\}$ and $Y$ respectively. 
The interpretation model $g$ is a function to explain the reason of output of machine learning model $\varphi$, 
which will be defined later and can quantify the contribution of feature $X$ to the output $Y$. 

Here, the feature contribution is a measure to evaluate the level how $X$ affects $Y$ in model $\varphi$. 
For example, SHAP is an interpretation method based on game theory \citep{lundberg2017unified}, 
which explains the predictions by computing the contributions of individual feature values. 
Its core idea is to consider each feature value as a player and the model's prediction as the success condition of a game. 
Then, it employs the Shapley value from game theory to calculate the contribution of each player to the success of the game. 

In order to unify the quantifying method w.r.t. the contribution of input $X$ to the output $Y$, we first present the following definition.

\begin{definition}[Interpretation List]
Given a dataset $D$, and $\varphi$ is a machine learning model trained from $D$, which has the input $X=\left\{x_{1}, x_{2}, \cdots, x_{n}\right\}$ and output $Y$. Given an interpretation model $g$ explains the output of model  by comparing the contribution of features to the output, which can be represented by the following list:
\begin{equation}
\left\{x_{i_{1}} \succ x_{i_{2}} \succ \cdots \succ x_{i_{n}}\right\}_{g}
\end{equation}
Here $x_{i_{a}} \succ x_{i_{b}}$  means that feature $x_{i_{a}}$ has larger contribution than  $x_{i_{b}}$ under the explanation $g$.

\end{definition}

In order to get a reasonable interpretation list, we need to find a unified mathematical description for the interpretation methods. Most interpretation methods, no matter the additive feature attribution method or its generalization, are used to explain a complex model’s output $f(x)$ , which try to approximate $f$ by defining an interpretation model $g$.

For the prediction $f(x)$ of an input $x$, we are trying to find a simplified input  $x^{\prime}$, 
and the relationship between  $x^{\prime}$ and the original input $x$ can be represented by the map $x=h_{x}\left(x^{\prime}\right)$ , with which the interpretation model satisfies $g\left(z^{\prime}\right) \rightarrow f\left(h_{x}\left(z^{\prime}\right)\right)$  
whenever $z^{\prime} \rightarrow x^{\prime}$ . So, we hope to find a unified paradigm to describe $g$ , 
as described in Scott and Lee's work \citep{lundberg2017unified}: 

\begin{example}
Additive feature attribution method is a linear function of the binary variables $z^{\prime} \in\{0,1\}^{M}$ :
\begin{equation}
g\left(z^{\prime}\right)=\phi_{0}+\sum\nolimits_{i=1}^{M} \phi_{i} z_{i}^{\prime}
\end{equation}
Here $M$ is the number of the simplified features. $\phi_{i} \in \mathbb{R}$, represents the effect of feature $x_{i}$ on the output $Y$.  $g\left(z^{\prime}\right)$ approximating the model output  $f(x)$ by adding the effects of all feature attributes.
\end{example}

Equation (2) can unify the six existing interpretation methods \citep{lundberg2017unified}. Here, the feature is simply regarded as a binary variable $z^{\prime} \in\{0,1\}^{M}$ , so that Equation (2) cannot describe some more complex interpretation methods. Thus we generalize the $z_{i}^{\prime}$  to a function, Equation (2) can describe more interpretation methods. Next, we construct a generalized definition of (2):
\begin{definition}
Let $g_{i}\left(x_{i}\right)$ is a function of feature $x_{i}$, 
$\phi_{i} \in \mathbb{R}$  represents the effect of each feature on the output,  
The principle of interpretation methods can be expressed as a linear combination of function,
\begin{equation}
    g(x)=\sum\nolimits_{i=1}^{n} \phi_{i} g_{i}\left(x_{i}\right)+\varepsilon
\end{equation}
where $\varepsilon$ represents the error between the interpretation model $g$ and the black box model $\varphi$.
\end{definition}

This function satisfying the following requirements: \\
i. \ $g(x) \neq 0$, which means that at least one feature can contribute to the model output. \\
ii. \ For simplification, the contribution of each feature $x_{i}$  can be fully described by $g_{i}\left(x_{i}\right)$,
and the interaction of any two features has no effect on the output.

\begin{remark}
In particular, Eq.(2) is the special case of (3), when $g_{i}\left(x_{i}\right)=\{0,1\}^{M}$; (3) is a generalized additive model, when $\phi_{i}=1$. \end{remark}
Compared with (2), the definition (3) can unify different interpretation methods into one framework.
The definition of (3) is necessary for the list (1). In (3), 
$g_{i}\left(x_{i}\right)$  can be arbitrary linear or nonlinear function, and it may be non-analytic in most cases. It is difficult for us to get the information of feature contribution from it directly, thus
we need to find an indicator to quantify the feature contribution. 
The indicator is $\phi_{i}$, which is the effect coefficient of the feature function  $g_{i}\left(x_{i}\right)$  and can also be understood as the weight. 
It means that we can use $\phi_{i}$ to compare the contribution of each feature to the output, 
resulting in interpretation list (1). To achieve this goal, we need to employ a series of mappings.

First, we map $g(x)$  in (3) to a vector (4) consisting of $\phi_{i}$. Only by extracting vector (4) from $g(x)$ can we further compare the contribution of features.

\begin{definition}
    Given a function space consisting of interpretation methods $\mathfrak{J}$  
    and a vector space $R^{n}$, $g(x)$  is an interpretation model of machine learning model $\varphi$. 
    The Coordinates Mapping  $\sigma: \mathfrak{J} \rightarrow \mathbb{R}^{n}$ extracts the coefficients of $g(x)$  and can be denoted as a vector:
\begin{equation}
    \sigma[g(x)]=\sigma\left[\sum\nolimits_{i=1}^{n} \phi_{i} g_{i}\left(x_{i}\right)+\varepsilon\right] \rightarrow\left(\phi_{1}, \phi_{2}, \cdots, \phi_{n}\right)
\end{equation}
\end{definition}
\noindent where $g_{1}\left(x_{1}\right), g_{2}\left(x_{2}\right), \cdots, g_{n}\left(x_{n}\right)$ are a set of basis vectors in the function space. The explanation model $g(x)$  is a linear combination of these basis vectors. $\sigma$  maps $g(x)$  to the coordinates $\left(\phi_{1}, \phi_{2}, \cdots, \phi_{n}\right)$  under the basis vectors $g_{1}\left(x_{1}\right), g_{2}\left(x_{2}\right), \cdots, g_{n}\left(x_{n}\right)$, which is an n-dimensional vector representing the effects of n features on the output.

Second, the elements of the vector in (4) are arranged in descending order, so that the contribution degree of each feature can be intuitively reflected and provide conditions for the next mapping.

\begin{definition}
    Given vector space $R^{n}$,the Descending Mapping $\tau: \mathbb{R}^{n} \rightarrow \mathbb{R}^{n}$  arranges the elements in the vector $\left(\phi_{1}, \phi_{2}, \cdots, \phi_{n}\right)$   in a descending order:
\begin{equation}
    \begin{aligned}\tau \circ \sigma[g(x)] & =\tau \circ \sigma\left[\sum\nolimits_{i=1}^{n} \phi_{i} g_{i}\left(x_{i}\right)+\varepsilon\right] \\& =\tau\left[\left(\phi_{1}, \phi_{2}, \cdots, \phi_{n}\right)\right] \\& =\left(\phi_{i_{1}}, \phi_{i_{2}}, \cdots, \phi_{i_{n}}\right)\end{aligned}
\end{equation}
\noindent where the vector $\left(\phi_{i_{1}}, \phi_{i_{2}}, \cdots, \phi_{i_{n}}\right)$ satisfies $\phi_{i_{1}}>\phi_{i_{2}}>\cdots>\phi_{i_{n}}$ . 
\end{definition}

Third, the elements of vector (5) are mapped to the corresponding features. In this way, the descending order vector of feature contribution can be obtained.
\begin{definition}
    Given a vector space $R^{n}$ and a feature space $\wp$ ,the Feature Mapping $\xi: \mathbb{R}^{n} \rightarrow \wp$ maps the elements of the vector $\left(\phi_{i_{1}}, \phi_{i_{2}}, \cdots, \phi_{i_{n}}\right)$ to the corresponding features:
\begin{equation}
    \xi \circ \tau \circ \sigma[g(x)]=\xi\left[\left(\phi_{i_{1}}, \phi_{i_{2}}, \cdots, \phi_{i_{n}}\right)\right]=\left(x_{i_{1}}, x_{i_{2}}, \cdots, x_{i_{n}}\right)
\end{equation}
\noindent where $\xi$ maps the effect $\phi_{i_{k}}$ to the feature $x_{i_{k}}$,$k \in(1,2, \cdots, n)$ . 
\end{definition}

Finally, we only need to replace the vector $\left(x_{i_{1}}, x_{i_{2}}, \cdots, x_{i_{n}}\right)$  with a notation to achieve our purpose. So we define a mapping that converts a vector into a special mark in this article.
\begin{definition}
    Given a feature space $\wp$  and its mapping space $\wp^{\prime}$, which consists of the interpretation lists.
    the Notation Conversion Mapping $\rho: \wp \rightarrow \wp^{\prime}$  casts the vector $\left(x_{i_{1}}, x_{i_{2}}, \cdots, x_{i_{n}}\right)$  to an interpretation list:
\begin{equation}
    \rho \circ \xi \circ \tau \circ \sigma[g(x)]=\rho\left(\left(x_{i_{1}}, x_{i_{2}}, \cdots, x_{i_{n}}\right)\right)=\left\{x_{i_{1}} \succ x_{i_{2}} \succ \cdots \succ x_{i_{n}}\right\}_{g}
\end{equation}
\end{definition}

Through the above composite mapping $\rho \circ \xi \circ \tau \circ \sigma$, the transformation from equation (3) to interpretation list (1) is realized. Next we will implement ensemble interpretation using interpretation list (1)

\begin{figure}[htbp]
\centering
\includegraphics[width=0.9\textwidth]{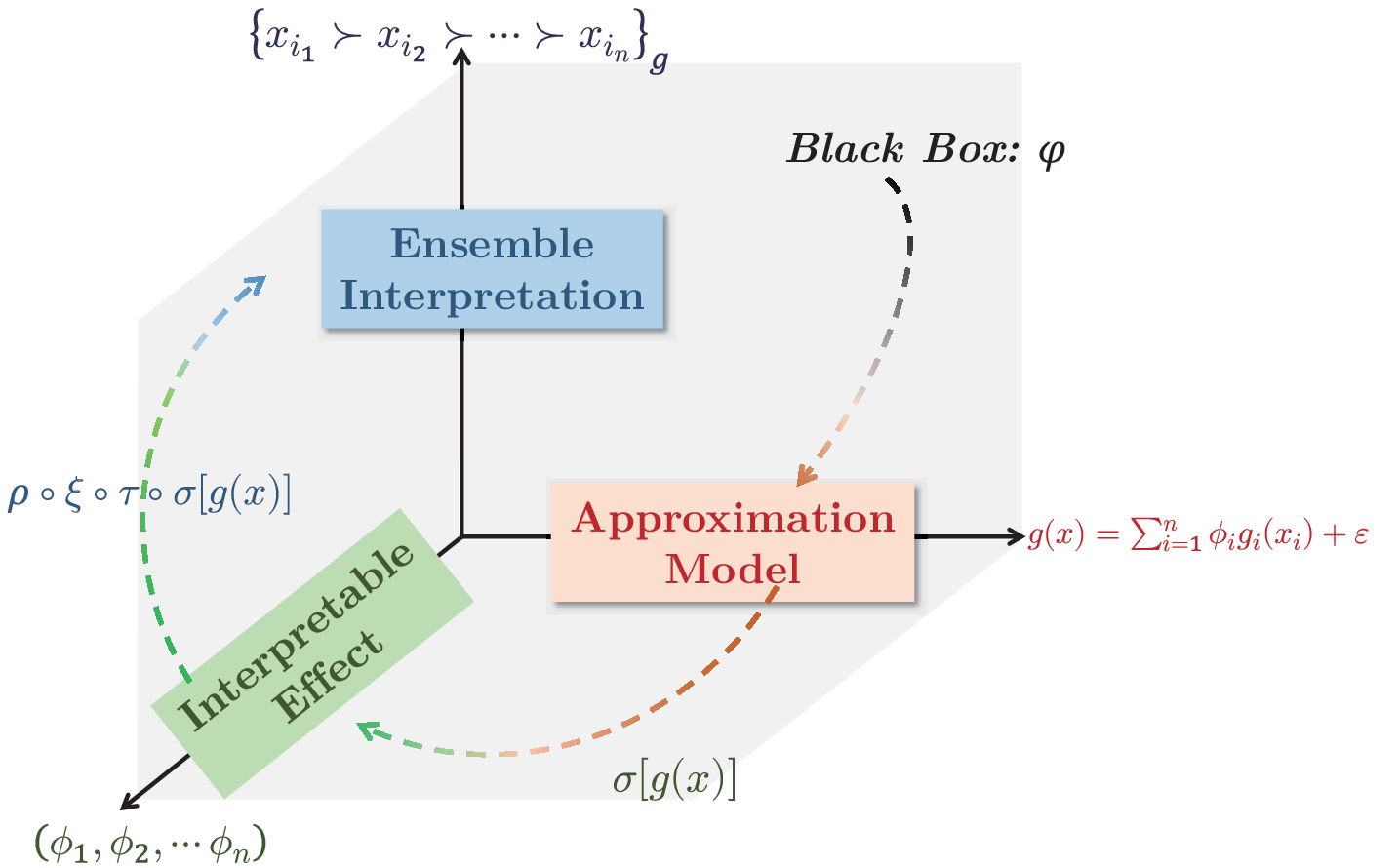}
\caption{The process of composite mapping. It shows that the interpretation model $g$ approximates the black box model $\varphi$ , composite mapping $\rho \circ \xi \circ \tau \circ \sigma$ realizes the transformation from interpretation model $g$ to interpretation list $\left\{x_{i_{1}} \succ x_{i_{2}} \succ \cdots \succ x_{i_{n}}\right\}_{g}$.}\label{fig1}
\end{figure}

The definition above regarding the effect $\phi_{i}$  is quite broad. In different interpretation  methods, the effect $\phi_{i}$  has different interpretations.
\begin{remark}
    In LIME \citep{ribeiro2016should}, it is to find $\phi_{i}$  by solving the following optimization problem:
\begin{equation}
    \xi=\underset{g \in \mathcal{G}}{\arg \min } L\left(f, g, \pi_{\dot{x}}\right)+\Omega(g)
\end{equation}

In DeepLIFT method \citep{shrikumar2017learning, shrikumar2016not}, the "summation-to-delta" property can be used to interpret effect $\phi_{i}$ . This property is defined as follows:
\begin{equation}
    \sum\nolimits_{i=1}^{n} C_{\Delta x_{i} \Delta o}=\Delta o
\end{equation}
where $o=f(x)$  represents the model output, $\Delta o=f(x)-f(r)$, $\Delta x_{i}=x_{i}-r_{i}$ and $r$ represents the reference input. In this context, $\phi_{i}$ corresponds to $C_{\Delta x_{i} \Delta o}$ , that is $\phi_{i}=C_{\Delta x_{i} \Delta o}$.

In classical Shapley Value methods \citep{datta2016algorithmic, lipovetsky2001analysis, vstrumbelj2014explaining}, $\phi_{i}$  is defined as:
\begin{equation}
    \phi_{i}=\sum_{S \subseteq F \backslash\{i\}} \frac{|S| !(|F|-|S|-1) !}{|F| !}\left[f_{S \cup\{i\}}\left(x_{S \cup\{i\}}\right)-f_{S}\left(x_{S}\right)\right]
\end{equation}
\noindent where $F$  is the set of all features,  $S$ is a subset of $F$ , $S \subseteq F$,  $f_{S \cup\{i\}}\left(x_{S \cup\{i\}}\right)$ represents a model trained on the subset $S \cup\{i\}$, $ f_{S}\left(x_{S}\right)$  represents a model trained on the subset $S$ . This method assigns an importance value to each feature, which indicates its impact on the model output.
\end{remark}

In addition to the aforementioned methods for interpreting  $\phi_{i}$ under different contexts, 
many other interpretation methods can also be found that correspond to its meaning. 
In summary, interpretation list (1) defined by  $\phi_{i}$  unifies most interpretation methods into the one paradigm.
Its essence lies in the impact of features on the output.

\section{Ensemble Interpretation}\label{sec4}
\subsection{The concept of explainer}\label{subsec4}
Each interpretation method satisfies Definition 2 and can generate an interpretation list through the composite mapping  $\rho \circ \xi \circ \tau \circ \sigma$ . The idea is to integrate the interpretation lists obtained by mapping different interpretation models $g^{i}$ to get an integrated interpretation list. We need to define the concept of explainer.

\begin{definition}
    The composite mapping $\rho \circ \xi \circ \tau \circ \sigma$  operates on an interpretation model  $g^{i}$ is called an explainer or a basic explainer, denoted by:
\begin{equation}
    E\left(g^{i}\right)=\rho \circ \xi \circ \tau \circ \sigma\left(g^{i}\right)
\end{equation}
where $E\left(g^{i}\right)$ returns an interpretation list of interpretation model denoted as
$E\left(g^{i}\right) \rightarrow l\left(g^{i}\right)=\left\{x_{i 1} \succ x_{i 2} \succ \cdots \succ x_{i n}\right\}_{g^{i}}$
\end{definition}

, which denotes the list returned by the $i$-th explainer. 
We call  $l\left(g^{i}\right)$ the interpretation list corresponding to $g^{i}$, 
where n  represents the number of input features, 
and  $x_{i j}(j=1,2, \cdots, n)$ is the $j$-th important feature ordered by the $i$-th explainer. 

Note that the symbol $\succ$  does not represent the conventional meaning of comparing numerical values, but rather indicates that the feature on the left of $\succ$  is more important for the model output than the feature on the right of $\succ$ . 

With the concept of base explainer, we naturally think of integrating these base explainers, the ensemble explainer obtained by integrating multiple basic explainers is represented as $E=\sum_{i=1}^{m} E\left(g^{i}\right)$,
where m is the number of explainers. 
Here $\sum_{i=1}^{m}$  represents the meaning of integration rather than summation. 
The result returned by $E$  is an integrated interpretation list 
$E \rightarrow l(E)=\left\{x_{i_{1}} \succ x_{i_{2}} \succ \cdots \succ x_{i_{n}}\right\}_{E}$.

In the process of integrating base explainers, these base explainers may be the same or different. We call the method of integrating the same base explainers as homogeneous interpretation, and call the method of integrating different explainers as heterogeneous interpretation.

\subsection{Integration Method}\label{subsec4}
Next we will present an integration method to ensemble the interpretation lists $l(E)$ of multiple explainers. 
The idea is to assign a score to each position in the interpretation list, 
then sum the scores obtained by each feature on each explainer to get the total score of each feature.
Finally, compare the total scores of each feature and sort them in descending order of scores to obtain the integrated interpretation list. 
Specifically, suppose there are  n features, for a certain explainer, the feature ranked first in the interpretation list obtains  n points, the feature ranked second obtains n-1  points, and so on, with the feature ranked last obtaining 1 point. Then, count the position of each feature on each interpretation list, add up the corresponding scores, and obtain the total score of each feature. Sort the features according to their total scores in descending order, and the result is the integrated interpretation list.

The algorithmic pseudo-code of ensemble interpretation is described as follows:
\renewcommand{\algorithmicrequire}{\textbf{Input:}}
\renewcommand{\algorithmicensure}{\textbf{Output:}}
\begin{algorithm}[h]
  \caption{Ensemble Interpretation}
  \label{alg::conjugateGradient}
  \begin{algorithmic}[1]
    \Require
     m base interpreter $E\left(g^{i}\right) \rightarrow l\left(g^{i}\right)=\left\{x_{i 1} \succ x_{i 2} \succ \cdots \succ x_{i n}\right\}_{g^{i}}$
    \Ensure
      ensemble interpreter $E \rightarrow l(E)=\left\{x_{i_{1}} \succ x_{i_{2}} \succ \cdots \succ x_{i_{n}}\right\}_{E}$
    \newline\Comment{Assign a score to each position in the interpretation list,$\bar{f}_{i}$ is the fraction assigned to the i-th position}
    \newline\Comment{$\hat{f}_{i k}$ represents the score of the $k$-th feature on the $i$-th explainer}
    \newline\Comment{$\hat{f}_{k}$ represents the total score of the $k$-th feature}
    \For {$j=n, n-1, \cdots, 1$}          
        \State $\bar{f}_{j}=j$;
    \EndFor
    \For{$i=1,2, \cdots, m$}
        \State $E\left(g^{i}\right) \rightarrow l\left(g^{i}\right)=\left\{x_{i 1} \succ x_{i 2} \succ \cdots \succ x_{i n}\right\}_{g^{i}}$
        \For{$k=1,2, \cdots, n$}
            \State $\hat{f}_{k}=\sum_{i=1}^{m} \hat{f}_{i k}$
        \EndFor
    \EndFor
    \State Put $\hat{f}_{k}$ corresponding features $x_{i_{k}}$ in descending order
    \State Return $\left\{x_{i_{1}} \succ x_{i_{2}} \succ \cdots \succ x_{i_{n}}\right\}_{E}$
  \end{algorithmic}
\end{algorithm}

\begin{figure}[h]
\centering
\includegraphics[width=0.9\textwidth]{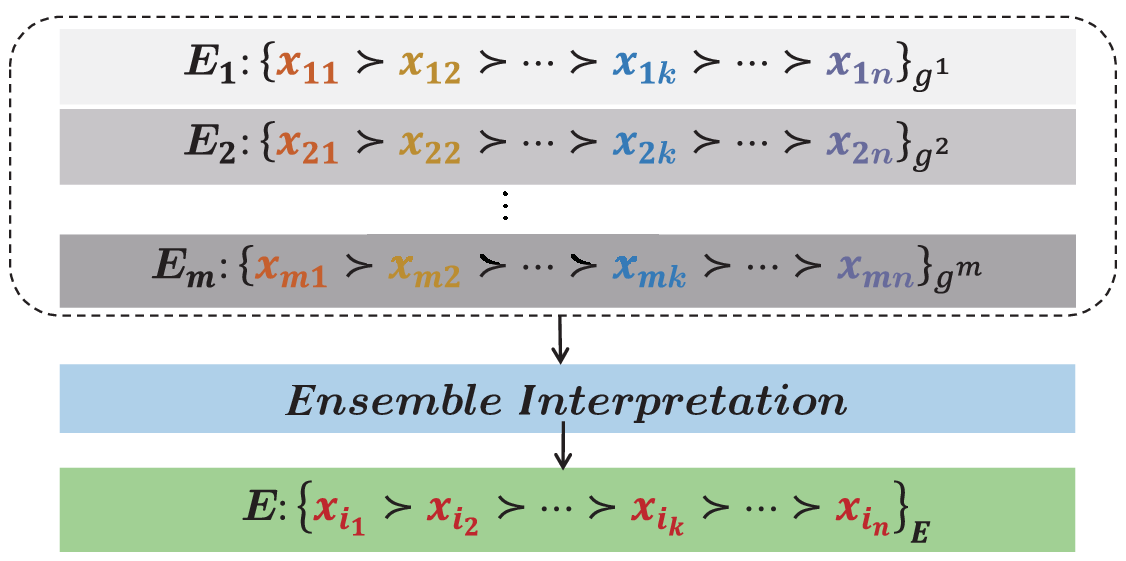}
\caption{Ensemble interpretation framework.Integrating a variety of interpretation lists to obtain an ensemble interpretation list.}\label{fig1}
\end{figure}

\subsection{Supervised evaluation of interpretation method}\label{subsec4}
It is hard to evaluate the performance of an interpretable method. 
In order to access on interpretation methods, 
we propose a evaluation index of interpretable methods based on the prior knowledge of tasks, 
which is called supervised evaluation of interpretation method.

For most machine learning tasks, human experts can get an interpretation list based on prior knowledge before explaining the model output.  

\begin{example}
A model $\varphi$ is now trained to predict whether students can pass the exam. Due to the limitations of the conditions, the currently collected data has only three features, including learning time, eating patterns (whether to eat on time), visual status (whether myopic). If we have completed the model training, we now need to explain the reasons for the output of the model. Before the explanation, we can know from experience that the model should make predictions based on the interpretation list:
\begin{equation}
\{\text {learning time} \succ \text {eating patterns} \succ \text {visual status}\}
\end{equation}
but if an interpretation method $g$  tells us that the model is based on the prediction made by
\begin{equation}
\left\{\text {visual status} \succ \text {eating patterns} \succ \text {learning time}\right\}_{g}
\end{equation}
then we can judge that this interpretation method is unreliable. Because we know that the learning time is obviously much more important than the other two features.
\end{example}
Therefore, our idea is to make full use of the prior knowledge of human beings to design an evaluation index that not only conforms to the law of human cognition, but also is concise, beautiful and reliable in mathematical expression. 

\begin{definition}
   For a machine learning task, the model $\varphi$ trained on the dataset $D$. 
   Let the input data sample be $x=\left(x_{1}, x_{2}, \cdots, x_{n}\right)$. 
   Before interpreting the model predictions, the human expert master the interpretation list 
   based on prior knowledge, which was called the reference interpretation label: 
\begin{equation}
    l=\left\{\bar{x}_{i_{1}} \succ \bar{x}_{i_{2}} \succ \cdots \succ \bar{x}_{i_{k}} \succ \cdots \succ \bar{x}_{i_{n}}\right\}
\end{equation}
\end{definition}
This interpretation label is similar to the label of data in supervised learning. it is the standard answer to the Interpreting outcome that conforms to human cognition. Since the interpretation label is not related to the interpretation method $g$, the symbol $\{\}$ has no subscript.

Suppose that the interpretation list obtained by an explainer 
$E(g)$ is $E(g) \rightarrow l(g)=\left\{x_{i_{1}} \succ x_{i_{2}} \succ \cdots \succ x_{i_{k}} \succ \cdots \succ x_{i_{n}}\right\}_{g}$.
Then, to quantify the gap between $L(g)$  and the interpretation label $l$, 
we define the following metric $\mathscr{L}_{score}$ \begin{equation}
    \mathscr{L}_{score}=\frac{\sum_{k=1}^{n} I_{k}}{n}
\end{equation}
where
\begin{equation}
    I_{k}=\left\{\begin{array}{l}1, x_{i k}=\bar{x}_{i k} \\0, x_{i k} \neq \bar{x}_{i k}\end{array}\right.
\end{equation}

The metric ranges from $[0,1]$, where 0 represents the worst case of the interpretation method, and 1 represents the most ideal case of the interpretation method. Generally speaking, the closer the $\mathscr{L}_{\text {score }}$ is to 1, the better the performance of the interpretation method and the closer it is to the logic of human thinking.

\section{Experiment}\label{sec5}

This experiment is conducted on two datasets: 
one is the red wine dataset from UCI with good data quality, 
and another is the natural gas engineering dataset with poor data quality. 
Considering the actual background of the two datasets and taking into account the prior knowledge, 
we provided interpretation labels to evaluate the performance of interpretation method. 
We applied two types of ensemble interpretation, homogeneous interpretation and heterogeneous interpretation to the two datasets respectively. 
Next, we will examine the performance of ensemble interpretation on these two datasets.

\subsection{Performance on Public Dataset}\label{subsec5}
\begin{table}[!ht] 
\centering 
\caption{Wine Quality Dataset}
\begin{tabular}{|c|c|c|c|r|l|} \hline 
\textbf{Dataset Name} & Wine Quality Data Set \\ \hline
\textbf{Type of task} & Classification \\ \hline
\textbf{Number of samples} &	178 \\ \hline
\textbf{Number of features} & 13 \\ \hline
\textbf{Number of Labels} & 3 \\ \hline
\end{tabular}
\end{table}

For simplicity and conciseness, we will write each feature of the wine dataset shown in Table 5. as an uppercase letter, as follows:
\begin{table}[!ht] 
\centering 
\label{Tab5}
\caption{The marked features of the wine quality dataset}
\begin{tabular}{|c|c|c|c|r|l|} \hline 
\textbf{Feature} & \textbf{Mark} \\ \hline
Alcohol & A \\ \hline
Malic acid & B \\ \hline
Ash & C \\ \hline
Alcalinity Of Ash & D \\ \hline
Magnesium & E \\ \hline
Total Phenols & F \\ \hline
Flavanoids & G \\ \hline
Nonflavanoid Phenols & H \\ \hline
Proanthocyanins & I \\ \hline
Color Intensity & J \\ \hline
Hue & K \\ \hline
Od280/Od315 Of Diluted Wines & L \\ \hline
Proline & M \\ \hline
\end{tabular}
\end{table}

According to the influence of each feature on the quality of wine shown in Table 5., 
the following interpretation label of wine data set are obtained through the 
investigation of experts in winemaking field :
\begin{equation}
    l=\{M \succ A \succ J \succ B \succ K \succ E \succ C \succ G \succ D \succ F \succ I \succ L \succ H\}\nonumber
\end{equation}

Decision Tree, SVM and Random forest etc. are the efficient algorithm, but its not suffciently interpretable. Here, we first use random forest to perform a classification task on the wine quality dataset, and then explain it. Next, we use the method of homogeneous interpretation and adopt the base explainer LIME to generate 11 explainers. The interpretation lists returned by these explainers are as follows:

\begin{table}[!ht] 
\centering 
\renewcommand\arraystretch{1.4}
\caption{Interpretation lists obtained by homogeneous base explainers}
\begin{tabular}{|c|c|c|c|r|l|} \hline 
LIME$_{1}$:$\left\{M \succ A \succ J \succ B \succ K \succ C \succ E \succ D \succ G \succ I \succ F \succ H \succ L\right\}_{g^{1}}$ \\ \hline
LIME$_{2}$:$\left\{M \succ A \succ J \succ B \succ E \succ K \succ C \succ G \succ I \succ F \succ D \succ L \succ H\right\}_{g^{2}}$  \\ \hline
LIME$_{3}$:$\left\{M \succ A \succ J \succ B \succ K \succ E \succ C \succ G \succ D \succ I \succ F \succ L \succ H\right\}_{g^{3}}$ \\ \hline
LIME$_{4}$:$\left\{M \succ A \succ J \succ B \succ K \succ E \succ F \succ C \succ D \succ G \succ I \succ H \succ L\right\}_{g^{4}}$\\ \hline
LIME$_{5}$:$\left\{M \succ A \succ J \succ B \succ K \succ G \succ E \succ L \succ C \succ F \succ D \succ I \succ H\right\}_{g^{5}}$\\ \hline
LIME$_{6}$:$\left\{M \succ A \succ J \succ B \succ K \succ E \succ C \succ D \succ I \succ F \succ G \succ L \succ H\right\}_{g^{6}}$\\ \hline
LIME$_{7}$:$\left\{M \succ A \succ J \succ B \succ K \succ E \succ C \succ G \succ D \succ I \succ F \succ L \succ H\right\}_{g^{7}}$\\ \hline
LIME$_{8}$:$\left\{M \succ A \succ J \succ B \succ K \succ E \succ C \succ D \succ G \succ F \succ I \succ H \succ L\right\}_{g^{8}}$\\ \hline
LIME$_{9}$:$\left\{M \succ J \succ A \succ B \succ K \succ E \succ C \succ G \succ D \succ F \succ I \succ L \succ H\right\}_{g^{9}}$\\ \hline
LIME$_{10}$:$\left\{M \succ A \succ J \succ B \succ K \succ C \succ E \succ G \succ D \succ I \succ F \succ L \succ H\right\}_{g^{10}}$\\ \hline
LIME$_{11}$:$\left\{M \succ A \succ J \succ B \succ K \succ C \succ E \succ D \succ G \succ F \succ I \succ H \succ L\right\}_{g^{11}}$\\ \hline
\end{tabular}
\end{table}

Combining the 11 interpretation lists in Table 3 and using the integration methods, 
we can get the ensemble interpretation list as follows:
\begin{equation}
    Ensemble: \left\{M \succ A \succ J \succ B \succ K \succ E \succ C \succ G \succ D \succ F \succ I \succ L \succ H\right\}_{E}\nonumber
\end{equation}

From the ensemble interpretation believes that feature M(Proline) has the greatest contribution to the wine quality, 
followed by feature A(Alcohol), 
and the feature holding the smallest contribution is H(Nonflavanoid Phenols).
Next, we use the metric $\mathscr{L}_{\text {score }}$ to evaluate the performance of single 
interpretation method and ensemble interpretation.

\begin{table}[!ht] 
\centering 
\caption{Performance comparison between interpretation method and ensemble interpretation for individual samples}
\begin{tabular}{|c|c|c|c|r|l|} \hline 
\textbf{Method} & $\mathscr{L}_{score}$ \\ \hline
LIME$_{1}$ & 0.3846 \\ \hline
LIME$_{2}$ & 0.6923 \\ \hline
LIME$_{3}$ & 0.8462 \\ \hline
LIME$_{4}$ & 0.6153 \\ \hline
LIME$_{5}$ & 0.5385 \\ \hline
LIME$_{6}$ & 0.6923 \\ \hline
LIME$_{7}$ & 0.8462 \\ \hline
LIME$_{8}$ & 0.6923 \\ \hline
LIME$_{9}$ & 0.8462 \\ \hline
LIME$_{10}$ & 0.6923 \\ \hline
LIME$_{11}$ & 0.5385 \\ \hline
\textbf{Ensemble (Our Method)} & \textbf{1.0000} \\ \hline
\end{tabular}
\end{table}

From the results in Table 4, we can see that compared to using a single interpretation method, 
the ensemble interpretation obtained exhibit consistent with human cognition. 
This also indirectly confirms that the performance of ensemble interpretation is better than that
of a single interpretation method.

\subsection{Performance on Natural Gas Dataset}\label{subsec5}
In this section, the natural gas engineering dataset is applied to conduct experiments and observe the performance of ensemble interpretation.
The dataset is sampled from a Coalbed Methane reservoir in China, Shanxi province,
which has been shown in Table 5 and 6 consisting of geological/engineering factors and 
the type of wells. Due to the complex working conditions and expensive annotation costs, the data is low quality

\begin{table}[!ht] 
\centering 
\caption{Natural Gas Dataset}
\begin{tabular}{|c|c|c|c|r|l|} \hline 
\textbf{Dataset Name} & Natural Gas Dataset \\ \hline
\textbf{Type of task} & Classification \\ \hline
\textbf{Number of samples} & 394 \\ \hline
\textbf{Number of features} & 20 \\ \hline
\textbf{Number of Labels} & 2 \\ \hline
\end{tabular}
\end{table}

For simplification, 
we will still use a letter to represent each feature, as in Table 6:
\begin{table}[!ht] 
\centering 
\caption{The features of the natural gas dataset are marked as capital letters}
\begin{tabular}{|c|c|c|c|r|l|} \hline 
\textbf{Feature} & \textbf{Marking} \\ \hline
Perforated section thickness & A \\ \hline
Total amount of fracturing fluid construction & B \\ \hline
Pre liquid volume & C \\ \hline
Sand carrying liquid volume & D \\ \hline
Pre liquid ratio & E \\ \hline
Liquid to liquid ratio for carrying sand & F \\ \hline
Total amount of proppant construction & G \\ \hline
Fracturing pressure & H \\ \hline
Maximum construction displacement & I \\ \hline
Maximum sand ratio & J \\ \hline
Average sand ratio & K \\ \hline
Sanding strength & L \\ \hline
Vertical stress & M \\ \hline
Maximum principal horizontal stress & N \\ \hline
Minimum horizontal principal stress & O \\ \hline
Gas saturation & P \\ \hline
Gas content & Q \\ \hline
Reservoir pressure & R \\ \hline
Temporary storage ratio & S \\ \hline
Permeability & T \\ \hline
\end{tabular}
\end{table}

Based on the analysis of the mission background by natural gas experts, the following interpretation label are obtained:
\begin{align}
    l &= \{Q \succ P \succ R \succ M \succ B \succ C \succ L \succ O \succ N \succ T \succ F \succ K \succ E \succ S \succ D \succ H \succ \nonumber \\
    & \ \ \ \ A \succ J \succ I \succ G\} \nonumber
\end{align}

Similarly, we use the data set to train the random forest, 
and now to explain the prediction of the model. 
Here, we will use heterogeneous interpretation methods to integrate seven commonly used interpretation methods, 
which are Shapley Additive Explanations (SHAP), Partial Dependence Plots (PDP), 
Accumulated Local Effects Plots (ALE), Permutation Feature Importance (PFI), Generative Additive Model (GAM), 
Global Surrogate Model (GSM), Feature Interaction (FI). The interpretation lists returned by each explainer is as follows:

\begin{table}[thbp] 
\centering 
\footnotesize
\renewcommand\arraystretch{1.4}
\caption{Interpretation lists obtained by heterogeneous base explainers.}
\begin{tabular}{|l|} \hline 
SHAP:$\left\{Q \succ R \succ P \succ L \succ M \succ S \succ T \succ N \succ B \succ O \succ I \succ K \succ F \succ C \succ A \succ H \succ D \succ G \succ E \succ J\right\}_{g^{1}}$ \\ \hline
PDP:$\left\{B \succ D \succ C \succ F \succ P \succ G \succ E \succ J \succ H \succ N \succ M \succ O \succ Q \succ L \succ K \succ I \succ A \succ R \succ S \succ T\right\}_{g^{2}}$ \\ \hline
PFI:$\left\{Q \succ R \succ M \succ P \succ L \succ K \succ O \succ S \succ D \succ N \succ E \succ B \succ T \succ D \succ J \succ I \succ H \succ G \succ C \succ A\right\}_{g^{3}}$ \\ \hline
GAD:$\left\{A \succ H \succ P \succ B \succ K \succ C \succ D \succ M \succ N \succ O \succ Q \succ R \succ S \succ T \succ F \succ G \succ E \succ I \succ J \succ L\right\}_{g^{4}}$ \\ \hline
GSD:$\left\{Q \succ L \succ R \succ C \succ E \succ P \succ T \succ A \succ K \succ M \succ S \succ I \succ G \succ F \succ O \succ B \succ D \succ H \succ N \succ J\right\}_{g^{5}}$ \\ \hline
FI:$\left\{T \succ Q \succ R \succ N \succ O \succ P \succ E \succ F \succ M \succ C \succ S \succ B \succ D \succ L \succ H \succ J \succ K \succ G \succ I \succ A\right\}_{g^{6}}$ \\ \hline
ALE:$\left\{Q \succ R \succ L \succ M \succ O \succ C \succ N \succ H \succ T \succ P \succ B \succ K \succ I \succ D \succ J \succ F \succ A \succ S \succ G \succ E\right\}_{g^{7}}$ \\ \hline
\end{tabular}
\end{table}

By combining the above 7 interpretation lists using integration method, the resulting ensemble interpretation list is as follows:
\begin{align}
    Ensemble &: \{Q \succ P \succ R \succ M \succ B \succ C \succ L \succ O \succ N \succ T \succ F \succ K \succ E \succ S \succ \nonumber \\ 
    & \ \ \ \ D \succ H \succ A \succ G \succ I \succ J\}_{E} \nonumber
\end{align}

Ensemble interpretation shows that feature Q (Gas content) has the greatest impact on 
distinguishing high-yield wells from low-yield wells, followed by feature P (Gas saturation), 
and the feature with the least impact is J (Maximum sand ratio). 
In order to test whether the interpretation results are consistent with human cognition, 
the $\mathscr{L}_{\text {score }}$ is used to evaluate the interpretation performance of 
single interpretation method  and ensemble interpretation.

\begin{table}[htbp] 
\centering 
\caption{Performance comparison between single interpretation method and ensemble interpretation}
\begin{tabular}{|c|c|c|c|r|l|} \hline 
\textbf{Method} & $\mathscr{L}_{score}$ \\ \hline
SHAP & 0.1500 \\ \hline
PDP & 0.0500 \\ \hline
PFI & 0.0500 \\ \hline
GAD & 0.0500 \\ \hline
GSD & 0.1000 \\ \hline
FI & 0.1000 \\ \hline
ALE & 0.2500 \\ \hline
\textbf{Ensemble (Our Method)} & \textbf{0.9000} \\ \hline
\end{tabular}
\end{table}

\begin{figure}[!htb]%
    \centering
    \includegraphics[width=0.9\textwidth]{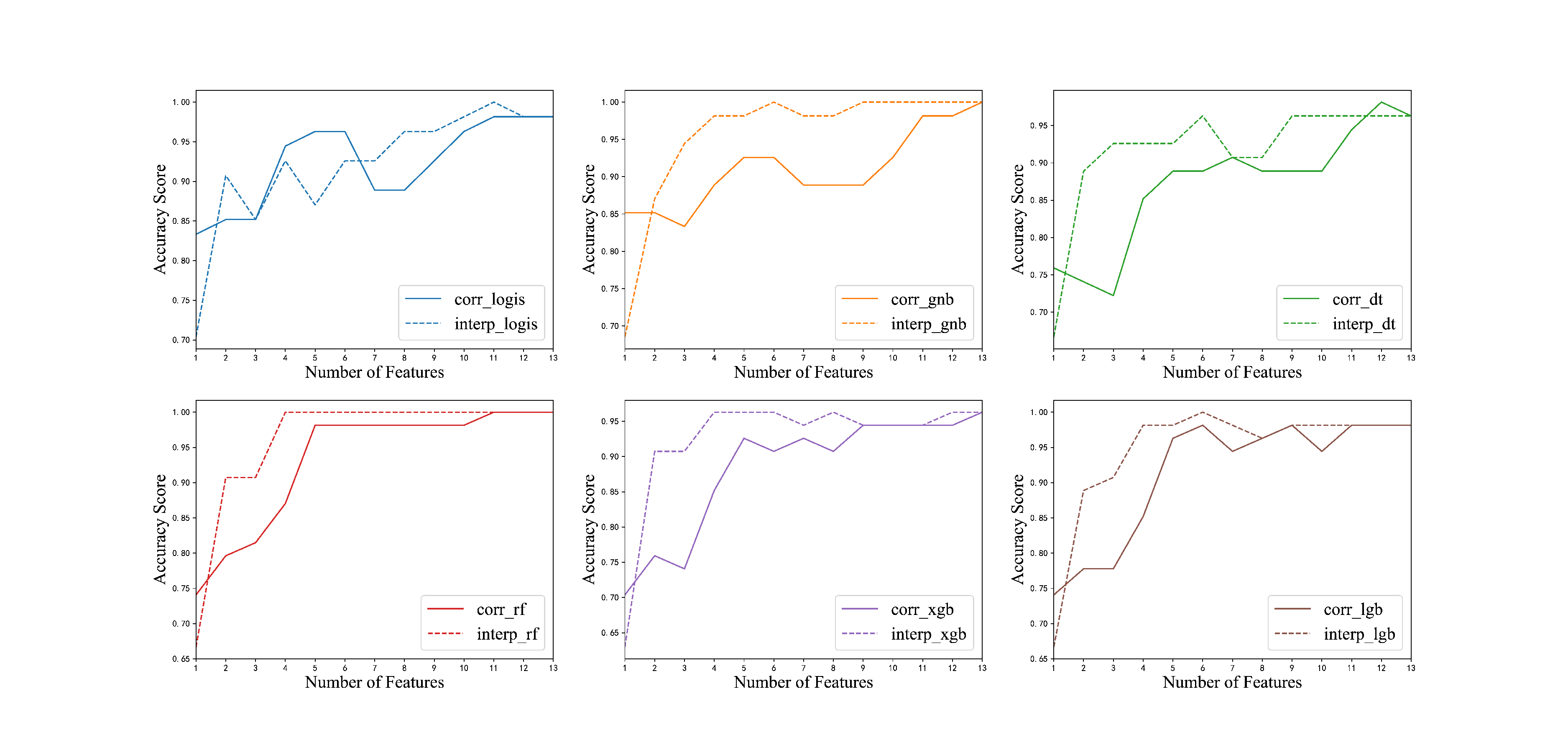}
    \caption{Performance comparison of ensemble interpretation and correlation on red wine dataset.}\label{fig1}
    \end{figure}
    
\begin{figure}[!htb]%
\centering
\includegraphics[width=0.9\textwidth]{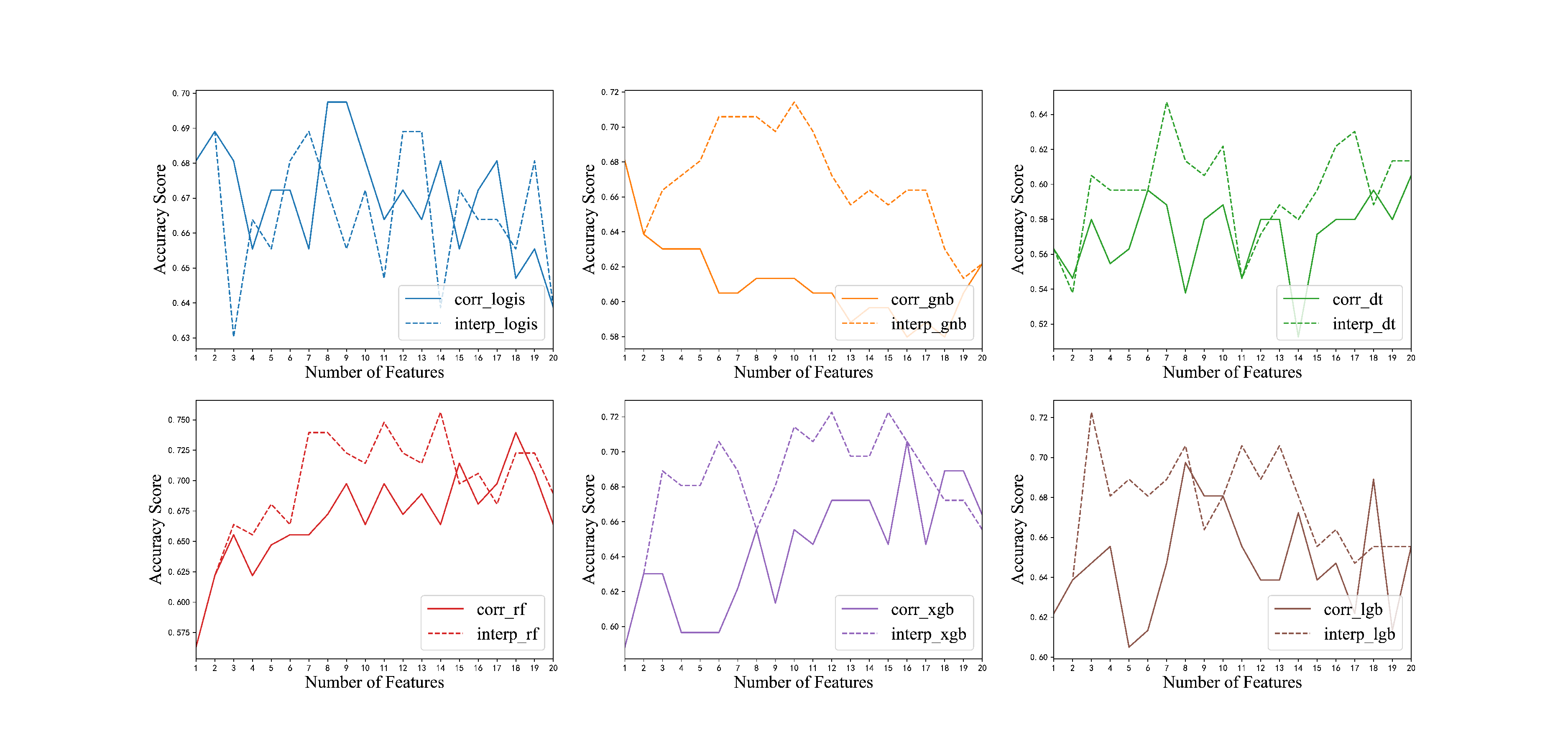}
\caption{Performance comparison of ensemble interpretation and correlation on natural gas dataset.}\label{fig1}
\end{figure}
It can be seen that on low quality dataset, the performance in Table 8 of other interpretation methods is not satisfying, 
but the ensemble interpretation still performs far better than the other individual interpretation method, 
showing better generalization and excellent interpretability, shown in Tbale 8 and Figure 3-4.
\subsection{Application in Improving Model Generalization Performance}\label{subsec5}
We have observed that ensemble interpretation performs better than other interpretation methods in terms of interpreting the model output, 
and the interpretation results obtained by the ensemble interpretation are more consistent with human cognitive experience. 
However, we are not satisfied the only interpretation and more
concerned that how to apply ensemble interpretation to improve the model's generalization ability. 
In this paper, we attempt to use the top $n$ feature in the interpretation list, 
where the selected features are more line with the human's cognition in engineering field and 
corresponding machine learning display the better generalization.
Features at the front of the interpretation list are retained, features at the back are discarded. 
The purpose is that features located later in the interpretation list 
do not contribute much to the model output and may affect the model's performance.
Thus, removing these poor quality features will help to improve the model's generalization performance. 
In order to compare the results, we also use correlation analysis for feature selection. 
We test the selected features on six classical machine learning algorithms, 
including logistic (logis), gaussian naive bayes (gnb), decision tree (dt), random forest (rf), XGBoost (xgb), Lightgbm (lgb). 
The results in Fig.3-4 are visualizations of the effects of the two feature selection methods on the improvement of model generalization performance.

In Fig 3 and 4, after feature selection with correlation analysis and ensemble interpretation, the performance of the model on the test set is compared. The corr\_model represents the performance of the model after feature selection using correlation analysis, and the interp\_model represents the performance of the model after feature selection using ensemble interpretation.
From the above result, it can be seen that the feature selection based on the ensemble interpretation significantly improves the generalization ability of various algorithms.

\section{Conclusion}\label{sec6}
In this paper, a guiding idea is presented to integrate multiple interpretation models, called ensemble interpretation. 
To achieve ensemble interpretation, a unified paradigm is defined for various interpretation methods, map it to different spaces, 
and obtain the interpretation list. 
Ensemble interpretation is realized by integrating multiple interpretation lists. 
In order to evaluate the performance of interpretation methods, 
a supervised evaluation of interpretation method is provided and
processed the evaluation index $\mathscr{L}_{score}$. 
In addition, ensemble interpretation can not only be applied to interpret model outputs, 
but also to implement feature selection using interpretation list to improve the generalization performance of models.

The proposed method can correct the deviation between the interpretation models and 
obtain a more stable interpretation result. 
However, there are still some limitations, 
including that there may be some interpretation models that cannot be mapped to interpretation list, 
and the acquisition of interpretation label will cause huge annotation cost. 

The future work direction of ensemble interpretation can be inspired by ensemble learning. 
By comparing the ideas of ensemble learning, 
we can know that the means to achieve ensemble learning are Boosting, Bagging, Stacking, etc. 
This means that the means to achieve ensemble interpretation can be various. 
The interpretation list defined in this paper is just one of many possible ways to implement ensemble interpretation, 
and we hope that future researchers can be inspired by the idea of ensemble interpretation to create more implementation methods. 
Furthermore, the unified paradigm we define expands the theoretical boundary of interpretable machine learning. 
This paradigm may also provide inspiration in other aspects for researchers.

\bibliography{sn-bibliography}%

\end{document}